\def\year{2017}\relax
\documentclass[letterpaper]{article} 
\usepackage{aaai17}  
\usepackage{times}  
\usepackage{helvet}  
\usepackage{courier}  
\usepackage{url}  
\usepackage{graphicx}  

\usepackage[utf8]{inputenc} 
\usepackage[T1]{fontenc}    
\usepackage{hyperref}       
\usepackage{url}            
\usepackage{booktabs}       
\usepackage{amsfonts}       
\usepackage{nicefrac}       
\usepackage{microtype}      
\usepackage{times}
\usepackage{algpseudocode}
\usepackage{comment}
\usepackage{algorithm}
\usepackage{soul}
\usepackage{amsmath}
\usepackage{graphicx} 
\usepackage{pbox}
\usepackage{subcaption}
\usepackage[dvipsnames]{xcolor}
\usepackage{float}

\newcommand{\xxnote}[3]{}
\ifx\hidenotes\undefined
  \usepackage{color}
  \renewcommand{\xxnote}[3]{\color{#2}{#1: #3}}
\fi

\newcommand{\San}[1]{{\xxnote{San}{red}{#1}}}

\begin{document}
%
\title{Understanding Convolutional Networks with \\ APPLE : Automatic Patch Pattern Labeling for Explanation}
\author{Sandeep Konam, Ian Quah, Stephanie Rosenthal, Manuela Veloso \\
    Carnegie Mellon University \\ 5000 Forbes Avenue, Pittsburgh, PA 15213 USA \\
    \{skonam,itq\}@andrew.cmu.edu,srosenthal@sei.cmu.edu,veloso@cs.cmu.edu}
\maketitle
\begin{abstract}
With the success of deep learning, recent efforts have been focused on analyzing how learned networks make their classifications. We are interested in analyzing the network output based on the network structure and information flow through the network layers.
We contribute an algorithm for 1) analyzing a deep network to find neurons that are ``important" in terms of the network classification outcome, and 2) automatically labeling the patches of the input image that activate these important neurons. We propose several measures of importance for neurons and demonstrate that our technique can be used to gain insight into, and explain how a network decomposes an image to make its final classification.
\end{abstract}

\section{Introduction}

Deep learning models have been shown to improve accuracy in a variety of application domains including image classification \cite{krizhevsky2012imagenet,szegedy2015going}, object detection (\cite{girshick2015fast,he2016deep}), and even robotics (e.g., localization \cite{yang2016pop}, navigation \cite{zhu2016target}, motion planning \cite{wulfmeier2016watch} and manipulation \cite{zhang2015towards}). Despite its success, there is still little insight into the internal operation and behavior of deep network models, or how they achieve such good performance \cite{zeiler2014visualizing}.

Many different algorithms have been proposed with the goal of explaining deep learning models, particularly for convolutional neural networks (CNNs) which analyze images. For example, Class Activation Maps (CAM) explain the output classification by visualizing a CNN's most discriminative pixels in the input image \cite{zhou2015learning}. Local Interpretable Model-agnostic Explanations (LIME) aims to learn a more interpretable model locally around the prediction and highlight the super-pixels with positive weight towards a specific class \cite{RibeiroSG16}. While these approaches have been successful in visualizing important pixels for classification, they do not explain how the network used those features to make the prediction.

To address the challenge of determining how a network uses image features, Fergus \& Zeiler \cite{zeiler2014visualizing} used deconvolutional networks \cite{zeiler2011adaptive} to visualize patterns that activate each neuron in the network (e.g., facial features such as the eyes or nose, other defining components of animals such as fur patterns, or even whole objects). However, their analysis of individual neurons and their image patch patterns was accomplished largely manually, which is impractical given the size of today's state-of-the-art networks as well as the number of images that are tested.

Building on this prior work, we contribute an algorithm to automatically label the features of an image that the network focuses on in order to explain why the network made its prediction. We accomplish this by analyzing the neurons that are most important to the output classification of an image as well as the patterns that activate those neurons. Our proposed approach, APPLE (Automatic Patch Pattern Labeling for Explanation), first analyzes the signal propagation through each layer of the network in order to find neurons that contribute highly to the signal in subsequent layers. Then, it deconvolves the important neurons at each layer to determine the parts or patches of the image that these neurons use as their input. Finally, our algorithm automatically labels the image patches with attributes that describe the object in the image using a separately-trained classifier.

We contribute several measures of neuron importance within a network and we demonstrate that our algorithm is able to use the measures to identify neurons that focus on important attributes of recognized objects. Beyond simply highlighting important pixels for visualization purposes, our image patch labels 
can be used to explain an image classification without requiring a human to decipher the image or manually probe the reasoning behind the prediction. On image classification tasks, we demonstrate that our APPLE algorithm reduces manual coding, finds important features of images, and automatically classifies those features for human interpretability.

\section{Related work}


We divide the prior research pertaining to understanding a CNN's predictions into two categories: weakly supervised localization and network structure analysis. 

In weakly supervised localization, the objective is to highlight the object features (pixels) within an image. For example, Class Activation Mapping (CAM) for CNNs use global average pooling to visualize the discriminative object parts detected by the CNN \cite{zhou2015learning}. Similar methods use other pooling techniques (e.g., global max pooling \cite{oquab2015object} and log-sum-exp pooling \cite{pinheiro2015image}), or reduce the network structure requirements that CAM imposes \cite{SelvarajuDVCPB16}. Local Interpretable Model-agnostic Explanations (LIME) is another weakly supervised localization technique \cite{RibeiroSG16}. In contrast to CAM which highlights discrimitive pixels, LIME finds a sparse linear approximation to the local decision boundary of a given black-box ML system including CNNs. Its visualization on an image allows a human operator to inspect how the classification depends locally on the most important input features. 
While these techniques use different measures for determining which pixels in the image are most important for classification, they do not analyze how the pixel features are propagated through the network to arrive at a prediction. 

We build on prior work that aims to understand how information propogates through a network (e.g., \cite{erhan2009visualizing,springenberg2014striving,mahendran2015understanding,zeiler2014visualizing}). For example, \cite{erhan2009visualizing} find the optimal stimulus for each neuron by performing gradient descent in image space to maximize the neuron’s activation. Our work uses deconvolutional networks, which are used to visualize which patterns activate each neuron \cite{zeiler2011adaptive,zeiler2014visualizing}.  Although most of the network structural analyses, including \cite{zeiler2014visualizing}, provide insights into a CNN's classification at the neuron-level, they require human intervention to manually analyze the activations or the important image patches to interpret how a network made a prediction. This manual process does not scale as the network gets bigger or as the number of images to analyze increases.

There is another category of work related to neural caption generation, where models learn to generate textual justifications for classifications of the primary neural model. \cite{Hendricks2016} use an LSTM
caption generation model with a loss function that encourages class discriminative information to generate justifications for the image classification of a CNN. \cite{ParkHASDR16} produce both textual justification and a visual attention map. \cite{VedantamBMPC17} produce captions that are locally discriminative, in the context of other images. However, all of these works use natural language descriptions at a large scale, collecting which has a prohibitive cost, compared to our work which only uses labels corresponding to important attributes of the objects.


Our Automatic Patch Pattern Labeling for Explanation (APPLE) algorithm is built on top of Fergus \& Zeiler \cite{zeiler2014visualizing}, but does not require any manual probing to understand the image patches that activate individual neurons, and we eliminate the need to analyze all of the neurons. In particular, we focus only on analyzing neurons that are important to the network. By assessing the ranked image patches corresponding to important neurons, APPLE simultaneously localizes the object within the image while also automatically labeling those patches with object feature attributes (e.g., eyes, nose, paws as attributes of the object class polar bears). 

\section{Automatic Patch Pattern Labeling for Explanation (APPLE)}

\begin{figure*}[t]
\centering
\setlength\fboxsep{0pt}
\setlength\fboxrule{0.25pt}
\includegraphics[width=5.5in]{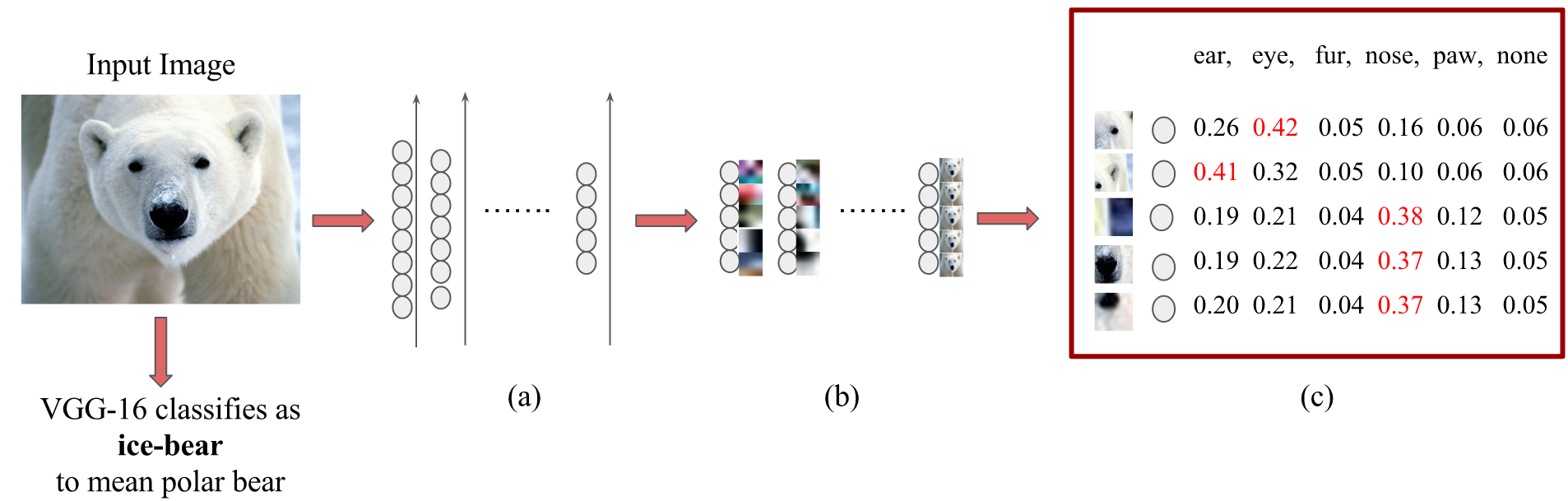}
\caption{The APPLE algorithm a) ranks neurons based on importance, b) identifies image patches corresponding to the top 5 neurons from (a), and then c) labels the patches using a secondary classifier to determine important object features.} 
\label{fig:pipeline}
\end{figure*}

Our goal is to explain the output of a CNN image classifier by ranking  the  neurons  based  on importance (i.e., most contribution to the final classification) and automatically labeling their corresponding patches of the image. For example, when classifying images of polar bears, some neurons within the network are activated more than others and it is likely that the most active neurons are detecting important attributes of polar bears (e.g., their eyes, nose, or paws). We produce a secondary classifier which takes as input the patches which correspond to the important or most active neurons and automatically labels them based on a predefined list of bear attributes. The combination of the important image patches and corresponding predicted attribute labels of those patches provide a qualitative understanding of what the CNN is using to classify polar bears based on what appears in its important regions in the image. 

In order to accomplish this goal, we propose our APPLE algorithm that:

\begin{enumerate}
    \item finds high importance neurons within the CNN, 
    \item deconvolves the network to determine the patch of the image that each important neuron looks at, and
    \item automatically labels those patches using a secondary classifier to determine object features that the patch contains.
\end{enumerate}

\subsection{High importance Neurons} \label{neurons}

We base our importance functions on the deep network structure. In that structure, a neuron $\eta$ in row $j'$ and column $k'$ in layer $l$ of a deep network takes input signal $X_{c', j', k'}$ on channel $c'$ (i.e., in images, r, g, and b each represent a channel). Neurons in layer $l$ are connected to layer $l$ + 1 by weights, $W$. A neuron's weight matrix $W_{\eta', \eta, m, n}$ weighs the activations of neuron $\eta$ in  layer $l$ and the rectangle of $m \times n$ neurons surrounding $\eta$ that connect to neuron $\eta'$ in layer $l + 1$. Convolving $W$ and $X$ produces output $Z$ where element $Z_{c, j, k}$ is the value of the output neuron within channel $c$ at row $j$ and column $k$. The activation $A_{c, j, k}$ is a function of $Z$, $\phi{(Z_{c, j, k})}$. Summarizing the definition formally, the forward propagation step as a CNN is described based on  \cite{Goodfellow-et-al-2016} as:

\begin{equation}\label{eq:second}
Z_{c, j, k} = \sum_{c, m, n} X_{c', j'+m-1 , k'+ n-1} W_{\eta', \eta, m, n}
\end{equation}

\begin{equation}\label{eq:third}
A_{c, j, k} = \phi{(Z_{c, j, k})}
\end{equation}

The propagation of information through the network can be considered a function of the post-activation output of a neuron, $A_{c, j, k}$ and the weights $W_{\eta', \eta, m, n}$. 

We propose four measures of importance based on the weights and activations of each neuron $(j,k)$ in layer $l$. Note that the weights and activations are a function of the matrix size $m \times n$ that surround each neuron, and therefore all of the importance measures of a neuron vary over indices $(j+\hat{m}, k+\hat{n})$ where $\hat{m}=(\frac{-m}{2}, \frac{m}{2})$ and $\hat{n}=(\frac{-n}{2}, \frac{n}{2})$.
\begin{itemize}
    \item \textbf{Activation Matrix Sum:} The sum of all values in the post-activation output $A_{c, j, k}$: 
    \begin{eqnarray}
    \sum_{\hat{m}, \hat{n}} A_{c, j + \hat{m}, k + \hat{n}}
    \end{eqnarray}
    \item \textbf{Activation Matrix Variance:} The variance of all values in the post-activation output $A_{c, j, k}$: 
    \begin{eqnarray}
    \sigma^{2}_{\hat{m}, \hat{n}} A_{c, j + \hat{m}, k + \hat{n}}
    \end{eqnarray}
    \item \textbf{Weight Matrix Sum:} The sum of all values in the weight matrix $W_{\eta', \eta, m, n}$: 
    \begin{eqnarray}
    \sum_{\hat{m}, \hat{n}} W_{\eta', \eta, m, n}[\hat{m}][\hat{n}]\cdot\varphi\text{, where } \varphi= 
    \begin{cases}
        1, & \text{if } A_{c, j, k}\geq 0\\
        0,              & \text{otherwise}
    \end{cases}
    \end{eqnarray}
    \item \textbf{Weight Matrix Variance:} The variance of all values in the weight matrix $W_{\eta', \eta, m, m}$: 
    \begin{eqnarray}
   \sigma^{2}_{\hat{m} , \hat{n}}\ \ W_{\eta', \eta, m, n}[\hat{m}][\hat{n}]\cdot\varphi\text{, where }\varphi= 
    \begin{cases}
        1, & \text{if } A_{c, j, k}\geq 0\\
        0,              & \text{otherwise}
    \end{cases}
    \end{eqnarray}
\end{itemize} 

Once the importance measure is computed for each neuron, they can be sorted and ranked to find the top neurons for each layer. These top neurons are used in the next step: patch extraction.

\subsection{Extraction of Patches corresponding to neuron}

Given the ranked neurons, we are interested in identifying the image patches that they convolve (Figure~\ref{fig:pipeline}b). The APPLE algorithm determines the image patches by deconvolving the network using a multi-layered Deconvolutional Network (deconvnet) as in Zeiler and Fergus (\cite{zeiler2014visualizing}). A deconvnet can be thought of as a convnet (CNN) model that uses the same components (filtering, pooling) but in reverse, so instead of mapping pixels to features does the opposite. 

To examine a given convnet activation, APPLE sets all other activations in the layer to zero and pass the feature maps as input to the attached deconvnet layer. Then APPLE successively (i) unpools, (ii) rectifies and (iii) filters to reconstruct the activity in the layer beneath that gave rise to the chosen activation. This is then repeated until the input pixel space, referred to as a patch is reached.

\subsection{Patch classifier}

In order to label the object attributes in each high importance patch, we construct and train a secondary classifier. Given a set of object attributes as classifier labels (e.g., eyes, nose, ear, fur, and paws for polar bears), we crop image patches as training data for each of these labels. We also include a `none' label which represents our background scene and parts of the object that may be difficult to distinguish. 

Once the patch classifier is trained, it can be run on the important image patches in order to determine the attribute label. Because we use a multi-class classifier, APPLE ranks the likelihood of each label on each patch (Figure~\ref{fig:pipeline}c) to determine the most likely label. Compared to \cite{zeiler2014visualizing} which requires manual probing to understand patches that activate neurons, our patch classifier automatically determines the \textit{labels} that can be used to explain what important areas of the image the network focused on. 

\subsection{Putting it all together}

\begin{figure*}[t]

    \centering
    \begin{subfigure}[t]{0.45\textwidth}
        \centering
         \includegraphics[height=50mm]{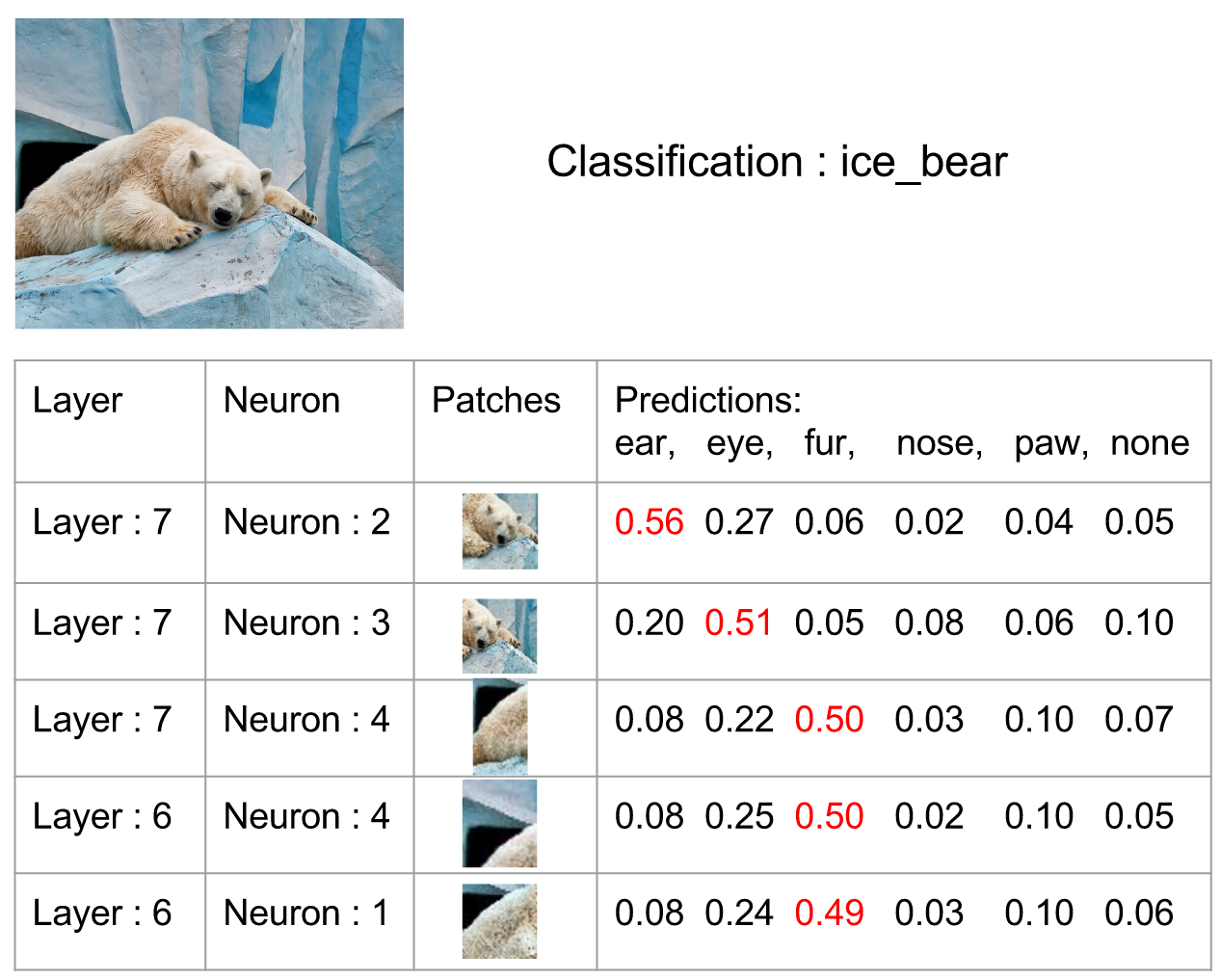}
        \caption{}
        \label{bear_example}
    \end{subfigure}%
     ~ 
    \begin{subfigure}[t]{0.45\textwidth}
        \centering
        \includegraphics[height=50mm]{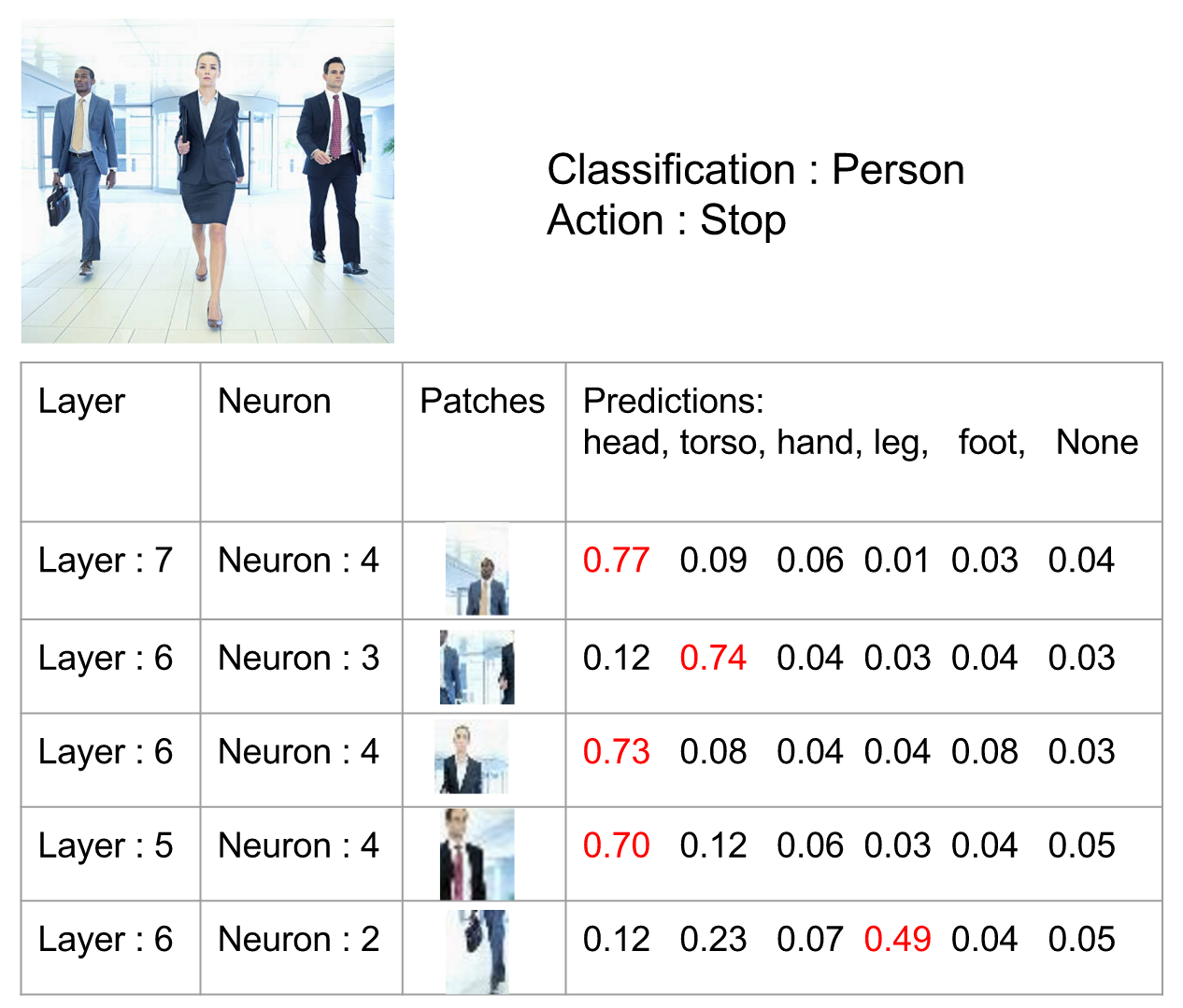}
         \caption{} 
          \label{person_example}
    \end{subfigure}

    \caption{APPLE sorts the labeled patches by confidence to present to a human in order to explain the CNN's image classification. Two example images are shown with their important patches selected using \textit{Activation matrix sum}.}
    \label{demo_all_examples}
\end{figure*}

Given an image (Figure~\ref{input_image:fig}) and a CNN model, our APPLE algorithm forward propagates the image to determine its classification. If APPLE has an attribute labeler for that class, it then automatically analyzes the CNN to find the top N most important neurons and constructs a list of image patch regions to label. Figure~\ref{patches:fig} shows the 15 image patches selected by the Activation Matrix Sum measure (top 5 neurons from the three layers - layers 5-7 - of interest) for polar bears. APPLE runs the patch classifier on the image patches to determine the most likely attribute labels (Figure~\ref{bear_example}). 
Labeled patches are then further sorted based on the maximum confidence. The important patches are a visual representation of the explanation of how the network determined the image's classification. The labels represent a semantic representation of the same explanation without requiring humans to interpret the image. 


\section{Experiments}

In order to demonstrate the ability of our APPLE algorithm to find important neurons and corresponding image patches and then automatically label them, we evaluated its use on two different datasets and three different object recognition tasks. In particular, we tested APPLE's ability to find the same 5 attributes of polar bears and dogs - eyes, ears, fur, nose, and paw - as well as its ability to find attributes of people - head, torso, hand, leg, and foot. We consider animal images particularly challenging as the background of the images often contains similar colors as the animals themselves, and because features that a person would consider to be strong distinguishers - black eyes and black nose - could easily be confused with each other and with rocks present in the environment. 

In this section, we describe our experimental setup using the VGG-16 classifier, our own trained patch classifer and the results of our experiments to evaluate the ability of our importance metrics to find regions of the object and the accuracy of our patch labels. 


\begin{figure}[t]
\centering
\includegraphics[width=3in]{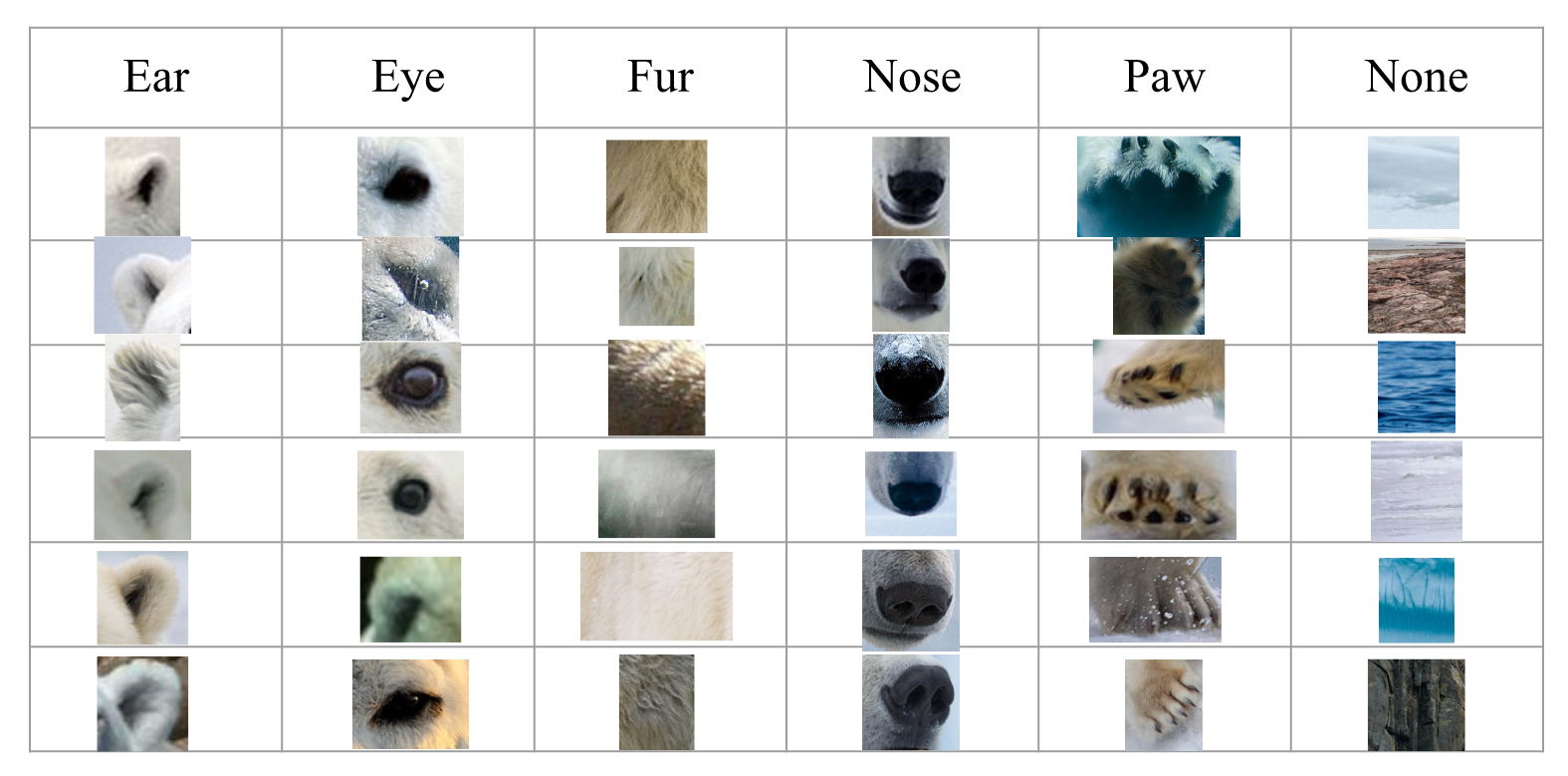}
\caption{Sample training data used to train the Patch classifier for the polar bear class}
\label{trainingdata}
\end{figure}

\subsection{CNN Classifier}

For the purpose of our experiment we chose the VGG-16 architecture \cite{Simonyan14c}  because 
it is a deep network with enough intermediate layers to gradually decrease the component granularity (to study increasing feature composition and its affect on final confidence). The VGG-16 architecture consists of 13 convolution layers followed by 3 dense layers, with max pooling after the 2$^{nd}$, 4$^{th}$, 7$^{th}$, 10$^{th}$ and 13$^{th}$ layer.  Despite the number of layers available for us to analyze, we only evaluated the image patches for neurons between layers 5 and 7 for two reasons. First, the image patches in the first few layers were too small to train a patch classifier for. Furthermore, results from Ranzato et al.\cite{Ranzato:2006:ELS:2976456.2976599} already show that the first layers learn stroke-detectors and Gabor-like filters. Additionally, image patches corresponding to the last few layers include a majority of the image and contain too many of the object attributes to accurately label just one.

In order to evaluate APPLE on different image classes and its importance functions on different weights within the VGG-16 architecture, we trained the VGG-16 architecture in two ways. To test our approach on polar bears and dogs, we used pre-trained weights for ImageNet\cite{russakovsky2015imagenet}  as provided in \cite{chollet2015keras}. To test our approach on people, we trained VGG-16 on the INRIA dataset \cite{dalal2005histograms}. 


 
 \subsection{Patch Classifier}

\begin{figure*}[t]

    \centering
    \begin{subfigure}[t]{0.4\textwidth}
        \centering
         \includegraphics[height=40mm]{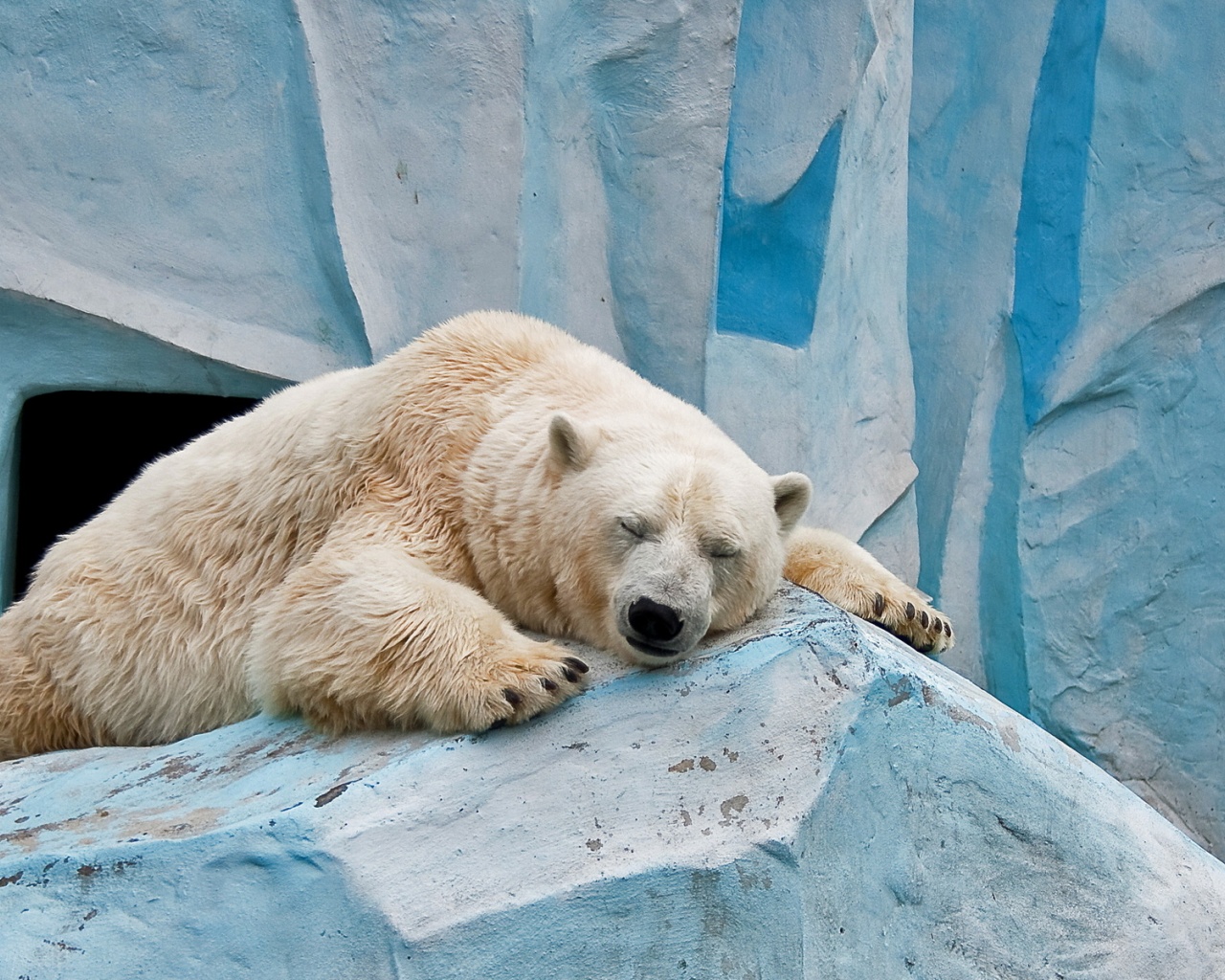}
        \caption{}
        \label{input_image:fig}
    \end{subfigure}%
     ~ 
    \begin{subfigure}[t]{0.52\textwidth}
        \centering
        \includegraphics[height=40mm]{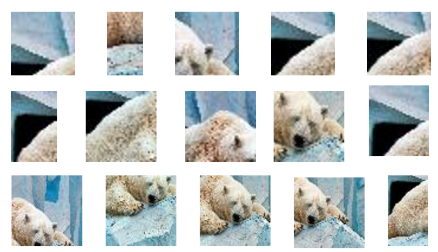}
         \caption{} 
          \label{patches:fig}
    \end{subfigure}
    ~ 
    \caption{(a): Input image. (b): High Importance Neuron Patches selected by APPLE algorithm.}
    \label{fig:patches_desc}
\end{figure*}

In order to train our patch classifier, we listed distinguishing attributes of polar bears and great pyranees or dogs - ears, eyes, nose, fur, and paws, and that of human beings - head, torso, hand, leg, and foot and manually cropped those important features from images in the Imagenet dataset \cite{russakovsky2015imagenet} and INRIA dataset \cite{dalal2005histograms} respectively. For the `none' class, we collected background patches from the same images. We had around 80 images representing each attribute, totalling the training data size to be 480 for each class (polar bear, dog, person). Each training image was 128x128 pixels. Figure~\ref{trainingdata} contains sample patches from our classifier dataset for polar bear. 

We used a multi-class Support Vector Classifier (AdaBoostSVM \cite{li2008adaboost}) with c=0.771, and $\gamma$ = 0.096, with an rbf kernel, determined by running K-fold cross validation on the patch data. The classifier obtained an average test set accuracy of 80\% when the data was split into 80\% training- 20\% test. SVMs were chosen for the task because of their ability to handle high dimensional complex data: an RBF kernel was used as described in \cite{li2008adaboost}. Our patch classifiers demonstrate the applicability of APPLE to a wide variety of image classes, and we note that we did not spend a large amount of time to acheive our attribute label results. It is possible to train a more accurate classifier on a larger dataset for even better results.

\begin{figure*}[t]
    \centering
    \begin{subfigure}[t]{0.3\textwidth}
        \centering
         \includegraphics[height=35mm]{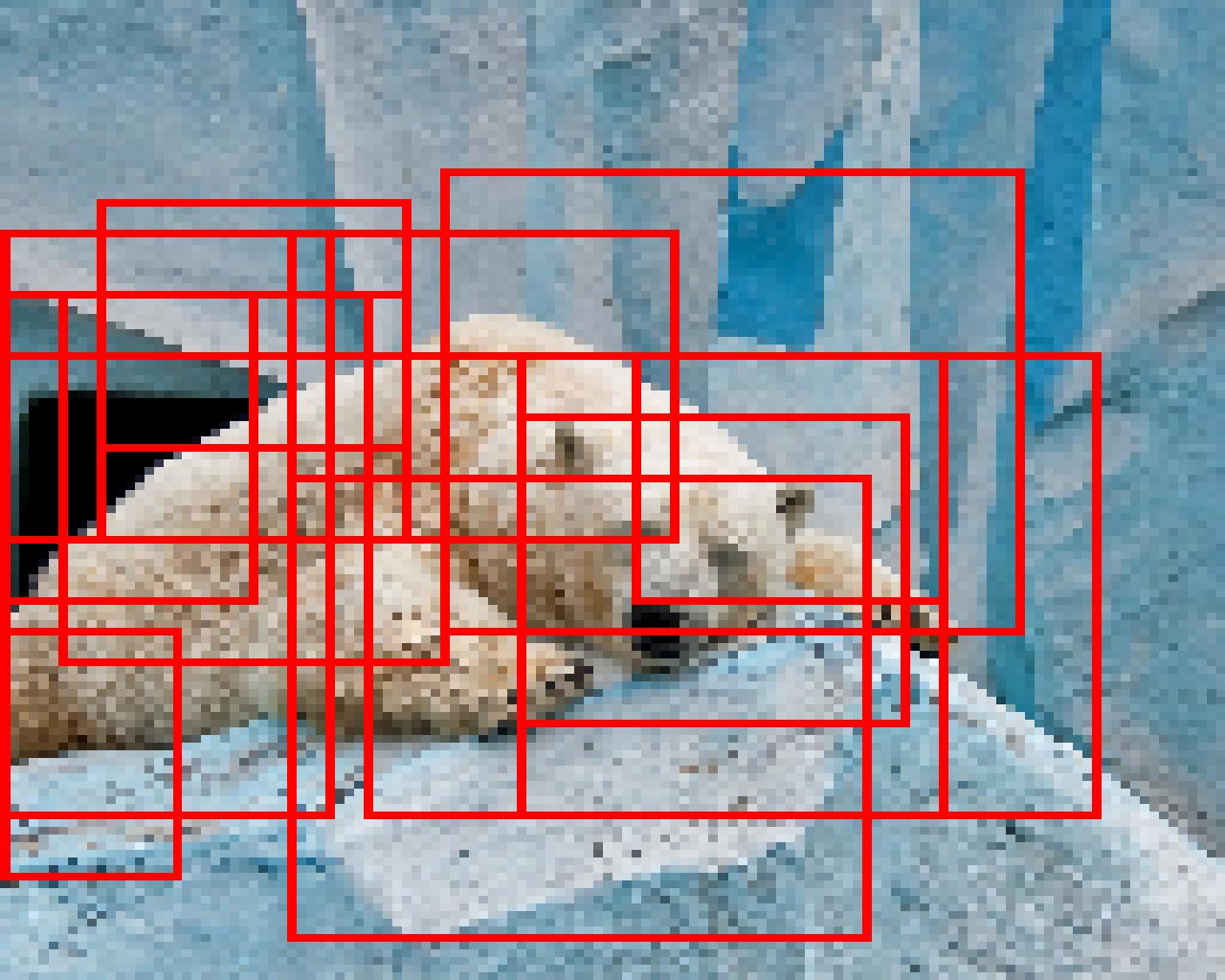}
        \caption{}
        \label{local_bear_boxes}
    \end{subfigure}%
     ~ 
    \begin{subfigure}[t]{0.3\textwidth}
        \centering
        \includegraphics[height=35mm]{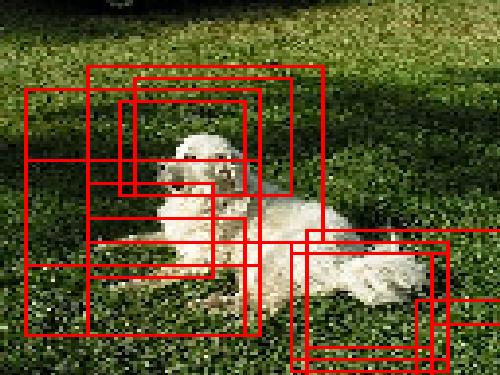}
         \caption{} 
          \label{local_dog_boxes}
    \end{subfigure}
    ~ 
    \begin{subfigure}[t]{0.3\textwidth}
        \centering
        \includegraphics[height=35mm]{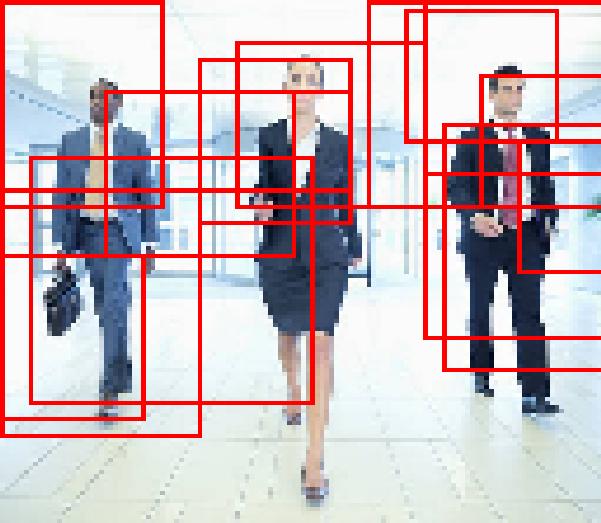}
         \caption{}
           \label{local_person_boxes}
    \end{subfigure}
     \begin{subfigure}[t]{0.3\textwidth}
        \centering
         \includegraphics[height=35mm]{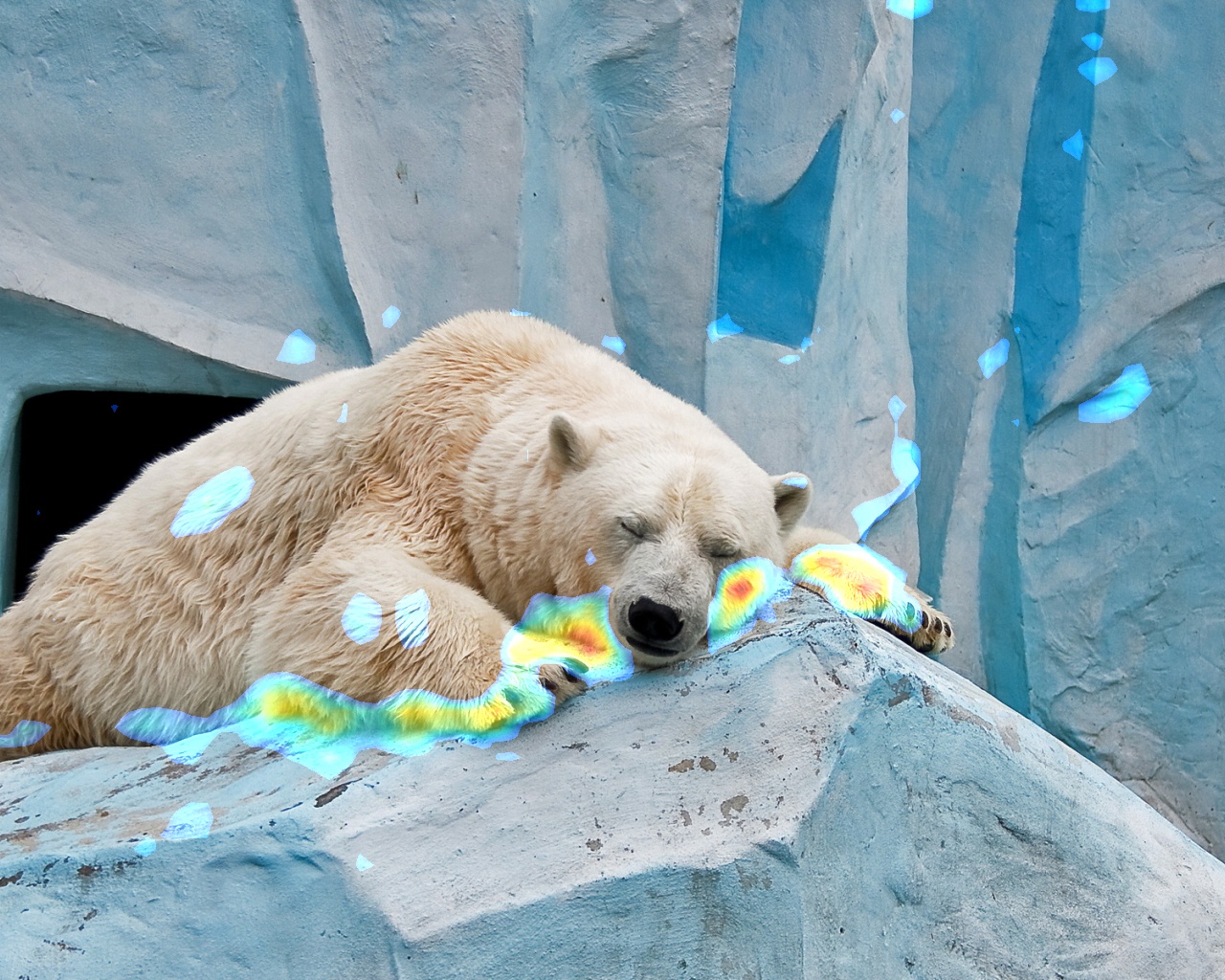}
        \caption{}
        \label{cam_bear_boxes}
    \end{subfigure}%
     ~ 
    \begin{subfigure}[t]{0.3\textwidth}
        \centering
        \includegraphics[height=35mm]{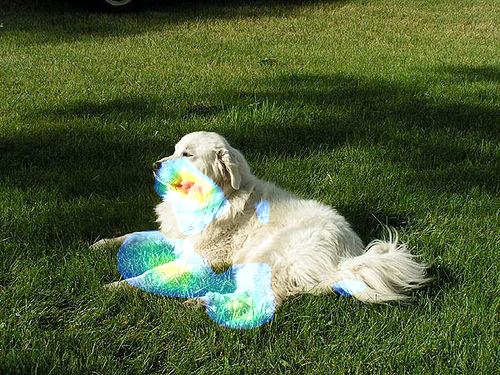}
         \caption{} 
          \label{cam_dog_boxes}
    \end{subfigure}
    ~ 
    \begin{subfigure}[t]{0.3\textwidth}
        \centering
        \includegraphics[height=35mm]{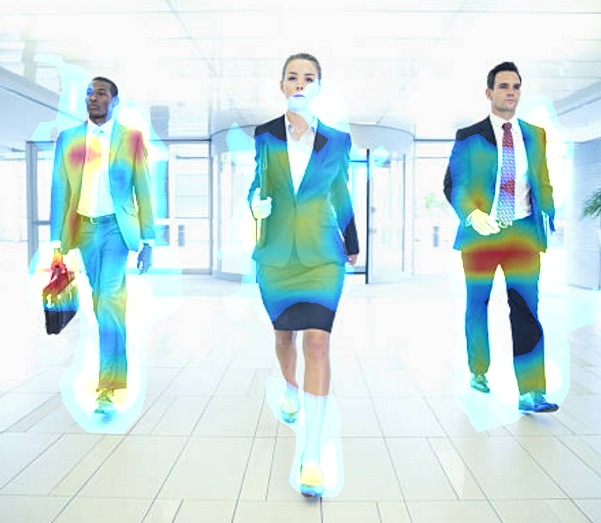}
         \caption{}
           \label{cam_person_boxes}
    \end{subfigure}
    \caption{APPLE and CAM each find important regions of images. In APPLE (a,b,c), red boxes indicate layers 5 - 7, and they encompass the entire object. On the CAM images (d,e,f), the heatmap visualizes its important pixels.}
    \label{fig:localization}
\end{figure*}

\subsection{Evaluation Setup}

Our experiments were conducted on 360 images, with 120 images per each class (polar bears, great pyrenees and persons). We selected the test images by searching for specific class on the Internet. 
Because our goal is to demonstrate the ability of APPLE to find important patches in any image and because the available labeled datasets for these classes is small, we chose to manually search for images of the classes in a variety of environments, poses, and conditions. The selection process involved criteria such as pose, number of classes visible in the image, lighting, and image resolution. By varying the different settings we could determine how different conditions influence the results of both our high importance neuron ranking method, as well as our patch classifier. We evaluate APPLE based on two measures of success: object localization and patch label precision.

\textbf{Object Localization Measure: } We first evaluated APPLE's ability to find image patches that were localized on the object of interest that the CNN is classifying (i.e., polar bears, dogs, and people in our experiments). \textit{Object localization} is measured as the ratio of number of patches containing pixels of the object (polar bear, dog, person) to the total number of patches. We analyzed three different types of objects to understand whether our importance functions are successful for different trained CNN classes, and compare against CAM's ability to find the important attributes of the object as well.

To do this, we outlined APPLE's important patches using the \textit{Activation matrix sum} measure on top of the original image. Figure~\ref{fig:localization} illustrates the localization abilities of our APPLE algorithm (red boxes indicate patches belonging to layers 5 - 7). We manually evaluated each image patch to determine whether it contains the object. For example, Figure \ref{patches:fig} shows 15 high importance patches picked by our heuristic. Each of these patches contains pixels belonging to polar bear, and hence they are all deemed as correct patches. In total, 15 important patches (top 5 patches across 3 layers) were evaluated as to whether they contained portions of the object (using our input image as reference). 

\textbf{Patch Label Precision Measure: } We evaluated patch classifier's precision at labeling the image patches. \textit{Patch label precision} is measured as the ratio of correctly classified patches to the total number of patches. A patch classification is considered correct if the top label output by the patch classifier matches our manually labeled ground-truth. 

\subsection{Results}

Table \ref{tab:patches} shows the precision of each of our evaluation measures on all 360 of the test images. We first note that our results were very similar for all four proposed importance metrics indicating that they are all successful and that any suggested measure could be used for the purpose of extracting neurons and thereby important patches. We compare the four-proposed metrics to a fifth metric which randomly picks 15 neurons (5 neurons across 3 layers). For clarity, we will report Activation Matrix Sum results which correspond to the image patches in Figure~\ref{fig:localization}. 

We first evaluated the abilty of our algorithm to localize objects in the image. As seen from Table \ref{tab:patches}, the four proposed importance metrics excel at object localization compared to the metric which randomly picks neurons. As seen from Figure~\ref{fig:localization}, most (93\%) of the red boxes are centered around the object, indicating that the important patches that APPLE finds are contributing highly to the classification. If we take into account the entire outline of our patches, they encompass the entire object, always more of the object than CAM does (Figure~\ref{fig:localization}d,e,f). This is significant because CAM requires modifications to the CNN architectures, whereas our approach works without any network modifications. Additionally, APPLE labels those important patches with object attributes, further automating the process. Because the patches are accurate for each of the three object classes and for different weights within the VGG-16 architecture, we conclude that APPLE's importance functions successfully and accurately find important patches across a variety of objects (polar bear, dog, and person) and datasets (ImageNet and INRIA).


Our next set of experiments focused on the accuracy of our patch labels. Table \ref{tab:patches} shows the patch label precision of the AdaBoost SVM on the important image patches for all 360 images. Our results show that we are able to label the image patches much more accurately than random guessing. However, the top-1 precision is less than 0.7 likely due to the choice of patch classifier and the quality of training data provided. For example, we only obtained 80 patches per label and the low-resolution of the images led to high confusion between the components (eyes and nose as well as fur and background (None)). We evaluated whether the correct label is in the top-2 predicted labels and found that the precision jumps to as high as 0.87. This result indicates that our trained patch labeler has high confusion with pairs of classes but more training data would help improve the precision further. It should be noted that results for the metric which randomly picks neurons are comparable to the four proposed metrics, which is unsurprising as `none' is one of the classes considered while training the patch classifier.  

Despite the challenges in building object attribute classifiers, our results demonstrate the patches identified as important by APPLE can be labeled accurately to help a human understand the CNN information propagation. This is the case even in the challenging tasks of labeling attributes of animals who's shapes and colors match the image backgrounds. We found that the results are the same across each of our three object classes.

\begin{table}
\caption {Evaluation of important patches averaged over all image classes.} \label{tab:patches}
\begin{center}
\begin{tabular}{ |p{2.8cm}|p{1.2cm}|p{1.4cm}|p{1.4cm}|}
 \hline
 Evaluation Measure & Object Localization & Label Precision (Top-1) & Label Precision (Top-2)\\
 \hline
Weight Sum & 0.91 & 0.610  & 0.810 \\
Weight Variance & 0.92  & 0.632   & 0.784\\
Activation Sum   & 0.93    & 0.697     & 0.832\\
Activation Variance &   0.91  &   0.641  &  0.871\\
\hline
Random & 0.54 & 0.614 & 0.782\\
 \hline
\end{tabular}
\end{center}
\end{table}


\section{Conclusion}
{
APPLE helps people gain a deeper understanding of what deep neural networks learn at the intermediate layers, and why they make the conclusions that they do. While prior work has focused on identifying important pixels that contribute to classification, little work has explored the impact of information propagation through the network.
 In this work, we contribute our algorithm, APPLE, to analyze the neuron-level information propagation and to facilitate the understanding of regions of interest. We contributed four different measures of neuron importance, such as sum and variance across the activation matrix and weight matrix of neurons. 

We demonstrated that in image classification tasks, our algorithm is able to use the measures to identify neurons within the CNN that focus on important attributes of the recognized object (i.e., body parts of animals). In particular, we demonstrated that all of APPLE's importance measures find regions of the images that contain the object of interest. We then used a patch classifier to label the attributes of the object, although it did confuse similar looking features of the bear. Currently, our approach uses one patch classifier for each class under consideration. At first glance, it might occur that our approach doesn't scale easily with increased number of classes, however it would be possible to automatically generate the list of features for each object (e.g., using web search) as well as crop attributes from images (e.g., using crowd-sourcing). Manual evaluation can also be automated by blocking out important patches and evaluating if the importance correlates with the change in class score. Future work could include constructing a universal patch dataset, improving patch classifier performance using a better classifier, more training data for a patch classifier and including context while classifying the patch. 

}

\bibliographystyle{aaai}
\bibliography{apple}

\end{document}